\documentclass{article} 
\usepackage[table]{xcolor}
\usepackage{collas2025_conference,times}
\usepackage{easyReview}

\usepackage{amsmath,amsfonts,bm}

\def\eqref#1{equation~\ref{#1}}

\def\1{\bm{1}}

\DeclareMathAlphabet{\mathsfit}{\encodingdefault}{\sfdefault}{m}{sl}
\SetMathAlphabet{\mathsfit}{bold}{\encodingdefault}{\sfdefault}{bx}{n}

\usepackage[T1]{fontenc}
\usepackage[margin=1in]{geometry}
\usepackage{wrapfig}
\usepackage{adjustbox}
\usepackage{booktabs}
\usepackage{hyperref}
\usepackage{multirow, graphicx}
\usepackage{algorithm}
\usepackage{algpseudocode}
\usepackage{listings}
\usepackage{amsmath}
\usepackage{capt-of}
\usepackage{titlesec}
\usepackage[shortlabels,inline]{enumitem}
\setlist[itemize]{noitemsep,leftmargin=*,topsep=0em}
\setlist[enumerate]{noitemsep,leftmargin=*,topsep=0em}
\usepackage{pifont}
\newcommand{\xmark}{\ding{55}}
\newcommand{\cmark}{\ding{51}}

\hypersetup{
    colorlinks=true,
    linkcolor=red,
    filecolor=magenta,
    urlcolor=blue,
    citecolor=purple,
    pdftitle={Overleaf Example},
    pdfpagemode=FullScreen,
    }

\title{Memory Storyboard: \\ Leveraging Temporal Segmentation for Streaming \\ Self-Supervised Learning from Egocentric Videos} 

\author{Yanlai Yang\\
New York University \\
\texttt{yy2694@nyu.edu} \\
\And Mengye Ren \\
New York University \\
\texttt{mengye@nyu.edu} \\
}

\collasfinalcopy 

\begin{document}
\maketitle

\begin{abstract}
Self-supervised learning holds the promise of learning good representations from real-world continuous uncurated data streams. However, most existing works in visual self-supervised learning focus on static images or artificial data streams. Towards exploring a more realistic learning substrate, we investigate streaming self-supervised learning from long-form real-world egocentric video streams. Inspired by the event segmentation mechanism in human perception and memory, we propose ``Memory Storyboard,'' a novel continual self-supervised learning framework that groups recent past frames into temporal segments for a more effective summarization of the past visual streams for memory replay. To accommodate efficient temporal segmentation, we propose a two-tier memory hierarchy: the recent past is stored in a short-term memory, where the storyboard temporal segments are produced and then transferred to a long-term memory. Experiments on two real-world egocentric video datasets show that contrastive learning objectives on top of storyboard frames result in semantically meaningful representations that outperform those produced by state-of-the-art unsupervised continual learning methods.
\end{abstract}

\section{Introduction}
\looseness=-10000
Humans are capable of learning continuously from a stream of unlabeled and uncurated perceptual inputs, such as video data, without needing to iterate through multiple exposures or epochs. Since early infancy, humans have accumulated knowledge about the world through a continuous flow of raw visual observations. This capability contrasts sharply with the training paradigm of current methods in self-supervised learning (SSL)~\citep{chen2020simple,grill2020bootstrap,chen2021exploring,caron2020unsupervised,bardes2021vicreg,he2022masked,assran2023self,he2020momentum}. Despite making significant strides in learning from large unlabeled datasets, these approaches still predominantly rely on static and curated image datasets, such as ImageNet~\citep{deng2009imagenet}, and require multiple epochs of training for effective learning. This difference in paradigm raises a compelling question: how can we learn good visual representations in a streaming setting---learning from visual inputs in their original temporal order without cycling back?

\begin{figure}[t]
\centering
\includegraphics[width=0.84\textwidth]{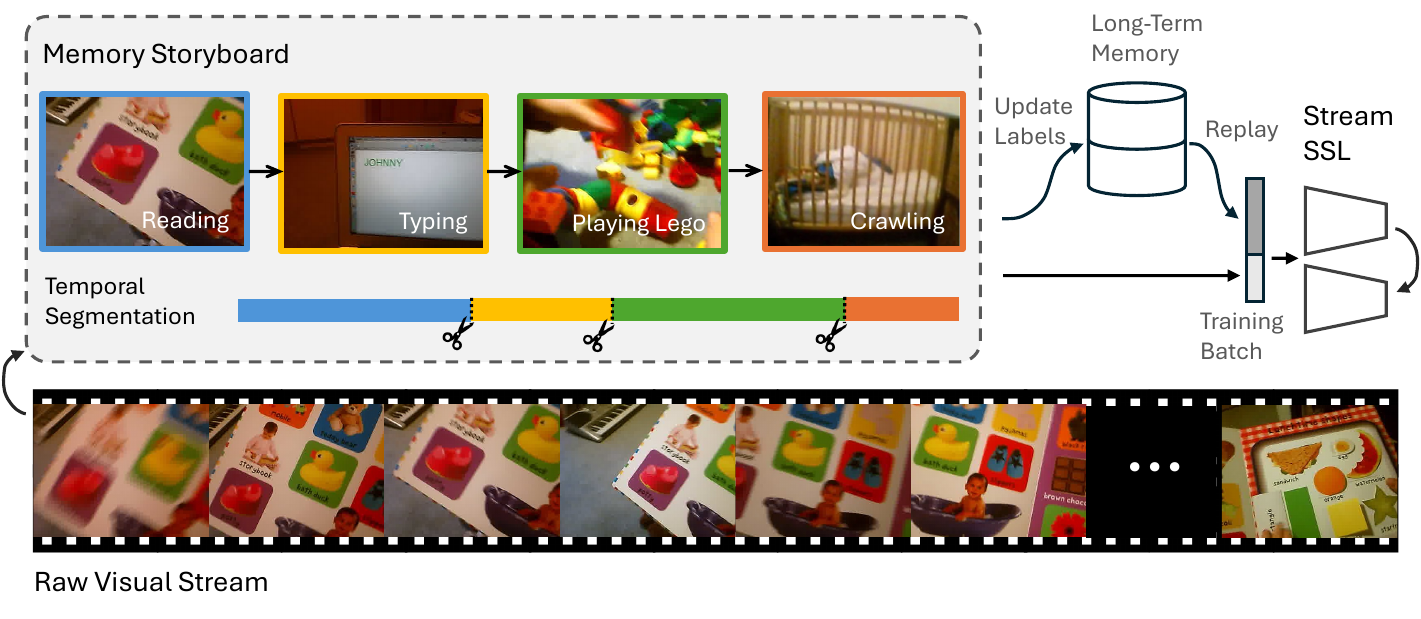}
\caption{\textbf{Memory Storyboard framework for streaming self-supervised learning (SSL) from egocentric videos.} Given a continuous stream of images from an egocentric video, Memory Storyboard effectively learns visual representations by clustering similar frames into temporal segments and updating their labels (text information for illustration purposes only) in the long-term memory buffer for replay. SSL involves contrastive learning at both the frame and temporal segment levels.}
\label{fig:teaser}
\vspace{0.1in}
\end{figure}
\looseness=-10000
Motivated by the differences in mechanisms between human learning and standard SSL, we aim to build learning algorithms that can efficiently learn visual representations and concepts from streaming video. One especially relevant mechanism in the human brain is event segmentation~\citep{newtson1977objective, zacks2001perceiving, yates2022neural}, where we spontaneously segment visual streams into hierarchically structured events and identify the event boundaries. Take your recent vacation trip as an example---you probably remember separate events and activities like exploring a city, dining at a local restaurant, or relaxing at the beach. The event segmentation mechanism helps us organize memories, recall specific moments, and summarize lengthened experiences~\citep{zacks2006event, zacks2007event}.

Drawing inspiration from the way we organize our memory in the brain, we introduce \textit{Memory Storyboard}, a novel approach for streaming self-supervised learning. Memory Storyboard features a temporal segmentation module, which groups video frames into semantically meaningful temporal segments, resembling the automatic event segmentation of human cognition. Through our temporal contrastive learning objective, these temporal segments effectively facilitate representation learning in streaming videos. To accommodate efficient temporal segmentation, we propose a two-tier hierarchical memory: temporal segmentation in the short-term memory is used to update the temporal class labels in the long-term memory, and a training batch consists of samples mixed from both memories. A high-level diagram of the algorithm is shown in Figure~\ref{fig:teaser}.

We conduct experiments on the SAYCam~\citep{sullivan2021saycam} and KrishnaCam~\citep{singh2016krishnacam} datasets of real-world egocentric videos. Memory Storyboard outperforms state-of-the-art unsupervised continual learning methods on downstream image classification and object detection tasks and significantly reduces the gap between streaming learning and the less flexible IID learning that requires persistent storage of the entire prior video data. We also experiment with different buffer sizes and batch sizes and offer insights into the optimal training batch composition under different memory constraints.

We summarize our contributions as follows:
\begin{enumerate}[1)]
    \item We introduce Memory Storyboard, a novel streaming SSL framework that features temporal segmentation and a two-tier memory hierarchy for efficient learning and temporal abstraction.
    \item We demonstrate that Memory Storyboard achieves state-of-the-art performance on downstream ImageNet~\citep{deng2009imagenet} and iNaturalist~\citep{van2018inaturalist} classification tasks when trained on real-world egocentric video datasets. Among all the streaming self-supervised learning methods we evaluated, Memory Storyboard is the only one that is competitive with or even outperforms IID training when trained on these datasets.
    \looseness=-10000
    \item We study the effects of training factors including label merging, subsampling rate, average segment length, memory buffer size, and training batch composition. These studies provide insight for more efficient streaming learning from videos. In particular, we explore the optimal composition ratio of the training batch from short-term vs. long-term memory, under different memory constraints. Larger batches from long-term memory improve performance when we can afford a large memory bank, while smaller batches can help prevent overfitting when we have a small memory bank.
\end{enumerate}
\section{Streaming SSL from Egocentric Videos}
\looseness=-10000
In streaming self-supervised learning, the goal is to learn useful visual representations from a continuous stream of inputs $(x_1, x_2, \dots)$. Similar to continual learning, we impose a memory budget so that storing the entire video would violate the constraint. Different from standard continual learning, there is no explicit notion of task, and the data distribution shift follows directly from the scene transitions of a video. The learner needs to make changes to the model as it sees new inputs, and finishes learning as soon as it receives the last input of the stream. The streaming setting is similar to Online Continual Learning~\citep{mai2021supervised, guo2022online, wei2023online}, but the focus here is primarily on streaming video frames instead of a fixed dataset of static images. We argue that streaming learning from sequential video frames enables better modeling of naturalistic scene transitions in real-world data streams because a stream of image collections often includes artificial class transitions.
\paragraph{Streaming Training Batches.} At each training step $t$, the model fetches a new batch of $b$ images $X_{t} = x_{tb:(t+1)b}$ from the video stream and updates its parameters upon receiving $X_{t}$. At the end of the video, we evaluate the final model checkpoint on various downstream tasks such as object classification and detection, which are fundamental tasks for visual scene understanding as they enable models to recognize and interpret the contents of complex environments.
\paragraph{Standard SSL Fails on Streaming Video.} Directly applying the SSL method sequentially on $X_{t}$ gives very poor performance~\citep{purushwalkam2022challenges,ren2021online}. This is not only due to catastrophic forgetting~\citep{mccloskey1989catastrophic} caused by the non-stationary distribution of visual features in the stream, but also due to the high temporal correlation of images in the stream (illustrated in Figure~\ref{fig:teaser}). This temporal correlation breaks the IID assumption held by common optimization algorithms like SGD or Adam~\citep{kingma2014adam}. For contrastive learning algorithms like SimCLR~\citep{chen2020simple}, the similarity across different frames in the same training batch would violate the assumption that each image is different.
\paragraph{Memory Replay.}
Similar to previous works~\citep{hu2021well, yu2023scale,purushwalkam2022challenges}, we use a replay buffer $M$ with finite size $|M|$ to mitigate these issues. The model can store some of the fetched images in the replay buffer, and use both samples from the replay buffer and the new frames to form a training batch of size $B$. By sampling from the replay buffer we de-correlate the frames in the training batch and at the same time reduce the distribution shift between training batches.
\paragraph{Benefits of Streaming SSL over Other Settings.}
Compared to the traditional self-supervised learning setting, where all the frames are shuffled and uniformly sampled for each batch (we refer to this as "IID learning" in the text below), streaming SSL allows embodied agents to learn good visual representations from natural, uncurated video streams. It also involves less computation delay and less memory storage. For instance, a robot in a new environment can continuously adapt the visual representations from its own egocentric video feed without any human curation.
\section{Related Work}
\label{sec:related}
In this section, we discuss the most relevant prior works. Please refer to Appendix~\ref{sec:supp_related} for additional related work.
\paragraph{Unsupervised Continual Learning.}
\looseness=-10000
Unsupervised Continual Learning (UCL)~\citep{rao2019continual, smith2019unsupervised, madaan2021representational, fini2022self, gomez2022continually, gomez2024plasticity, cheng2023contrastive, zhang2024integrating} aims at learning a good representation through an unlabeled non-stationary data stream. Existing works in UCL often assume that the data stream is composed of a series of episodes and a stationary data distribution within each episode. This is not as naturalistic and human-like as our streaming setting, where the data distribution changes continuously through the data stream, and each image appears in the data stream only once. Meanwhile, we showed that existing UCL methods are also effective in our streaming video setting, and can be used together with the supervised contrastive objective.
\paragraph{Streaming Learning from Videos.}
\looseness=-10000
While a number of recent papers have studied streaming learning from images~\citep{hayes2019memory, hayes2020lifelong, hayes2020remind, banerjee2021class} or IID self-supervised learning from video frames~\citep{venkataramanan2023imagenet, wang2024poodle}, limited works have investigated the problem of streaming learning from a continuous video stream. \citet{roady2020stream} introduces a benchmark for streaming classification and novelty detection from videos. \citet{zhuang2022well} benchmarks many self-supervised learning methods in real-time and life-long learning settings in streaming video, assuming infinite replay buffer size which is unrealistic. Most similar to our setup,~\citet{purushwalkam2022challenges} studies the task of continuous representation learning with a SimSiam objective~\citep{chen2021exploring} and proposes using a minimum-redundancy replay buffer. Their work also belongs to the broader range of works that study replay buffer sampling strategies in continual learning~\citep{aljundi2019gradient, wiewel2021entropy, tiwari2022gcr, hacohen2024forgetting}. Our work extends these prior works by adopting a two-tier replay buffer and a temporal segmentation component. Also relevant to our work,~\citet{carreira2024learning} studies online learning from a continuous video stream using a pixel-to-pixel reconstruction loss for representation learning. Their findings on the effect of pre-training and different optimization schemes are orthogonal with the ones in our work. It is worth pointing out that their exploration mainly focuses on settings without data augmentation and replay, limiting the efficacy of their framework.
\section{Memory Storyboard}
\begin{wrapfigure}{r}{0.5\textwidth}
    \centering
    \includegraphics[width=\linewidth]{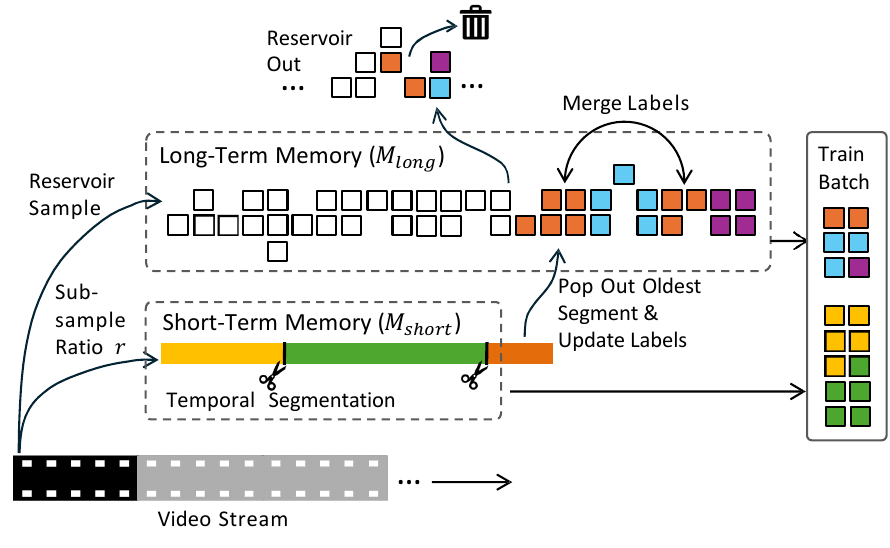}
    \caption{\textbf{Details of our two-tier memory in Memory Storyboard.} Long-term memory is updated with reservoir sampling and short-term memory with first-in-first-out (FIFO). Temporal segmentation is applied on the short-term memory, which then updates the labels of corresponding images in the long-term memory.}
    \label{fig:detail}
    \vspace{-0.1in}
\end{wrapfigure}

\looseness=-10000
We present Memory Storyboard, an effective method for streaming SSL from egocentric videos. Memory Storyboard includes a temporal segmentation module and a two-tier memory hierarchy. It combines a standard self-supervised contrastive loss with a temporal contrastive objective that leverages the temporal class labels produced by the temporal segmentation module. Figure~\ref{fig:detail} illustrates the details of our method. The overall data processing and training procedure is summarized in Algorithm~\ref{alg:memstoryboard}.
\paragraph{Temporal Segmentation.} We describe our temporal segmentation algorithm as follows. Similar to~\citet{potapov2014category}, we are given a down-sampled video frame sequence of length $L$, with frames $x_1, x_2, \cdots, x_L$, and a feature extractor $f_\theta$. We aim to find change points $t_1, t_2, \cdots, t_{n-1}$ so that the video is divided into $n$ semantically-consistent segments $[x_1, x_{t_1}], [x_{t_1}, x_{t_2}], \cdots, [x_{t_{n-1}}, x_L]$. We also define $t_0 = 0$ and $t_n = L$. In this work, we determine the number of segments with $n = \frac{L}{T}$, where $T$ refers to the average segment length and is a hyper-parameter.

The optimization objective of our segmentation algorithm is to maximize the average within-class similarity, such that each temporal segment captures a coherent scene, i.e.
\begin{equation}
    \max_{t_1, t_2, \cdots, t_{n-1}} \sum_{i=2}^n \frac{1}{t_{i} - t_{i-1}} \sum_{j=t_{i-1}}^{t_{i}} \sum_{k=j}^{t_{i}} sim(x_j, x_k).
\end{equation}
where $sim(x_j, x_k)$ denotes the cosine similarity between the embeddings $f_\theta(x_j)$ and $f_\theta(x_k)$. 
We compute the approximate solution to this optimization problem with a greedy approach, as detailed in Algorithm~\ref{alg:temporalseg}. We adopt this simple temporal segmentation approach in order to get good segmentation results in the beginning when the encoder network does not provide good representations. We leave it to future work to investigate different temporal segmentation strategies.

\paragraph{Two-tier Memory Hierarchy.} Inspired by the Complementary Learning Systems (CLS) theory~\citep{mcclelland1995there, o2014complementary} of the human brain, we propose a two-tier memory hierarchy to accommodate efficient temporal segmentation. Shown in Figure~\ref{fig:detail}, the system includes a long-term memory $M_{long}$ updated with reservoir sampling~\citep{vitter1985random}, and a short-term memory storyboard $M_{short}$ updated with a first-in-first-out (FIFO) strategy. We store the temporal index and the temporal class of each frame along with the image in the memory. The short-term memory size $|M_{short}|$ is much smaller than the long-term memory size $|M_{long}|$, allowing efficient temporal segmentation of the recent past. The change points produced by the temporal segmentation component on $M_{short}$ are then used to update the temporal class labels in $M_{long}$.

To increase the horizon of the memory storyboard, we subsample the frames coming from the current stream before adding it to $M_{short}$. The subsampling also reduces the temporal correlation between the frames in the training batch sampled from  $M_{short}$. Compared to using a single replay buffer as memory, the two-tier memory hierarchy helps avoid overfitting the replay buffer and makes sure that the new frames are seen by the model.

\paragraph{Label Merging.} Same objects and scenes often repeat in egocentric video streams.
To efficiently share visual concept labels, we introduce here a label merging mechanism. When a new temporal segment is added to $M_{\text{long}}$, we compute the cosine similarity between its average frame embedding and those of existing segments. If the maximum similarity exceeds a threshold $\delta$, the new segment inherits the class label of the most similar segment. This mechanism is activated only after the first $C$ segments, as early-stage embeddings tend to be uniformly high in similarity. Formally, let $v_i$ denote the average embedding of segment $i$ in $M_{\text{long}}$, and $v_n$ for the new segment $n$. Define $j = \arg\max_i \text{sim}(v_i, v_n)$ and let $c_j$ be the label of segment $j$. Then, 
\[
c_n = 
\begin{cases}
c_j & \text{if } \text{sim}(v_j, v_n) > \delta \text{ and } n > C \\
\text{new label} & \text{otherwise}.
\end{cases}
\]
In practice, choosing a fixed threshold $\delta$ that generalizes across methods and datasets is challenging. To address this, rather than fixing $\delta$ manually, we define it dynamically based on a quantile threshold $\tau \in (0, 1)$. Specifically, we set $\delta$ as the $\tau$-quantile of all off-diagonal values in the similarity matrix.

\paragraph{Temporal Contrastive Loss.} To effectively utilize the temporal class labels for representation learning, we adopt the supervised contrastive (SupCon) loss~\citep{khosla2020supervised}, which takes the samples with the same temporal class label in a batch as positives and contrasts them from the remainder of the batch. 
Let $f_{proj}$ be a projector network. For a batch of images with size $B$, we take two random augmentations of each image to get an augmented batch $\tilde{x}_1, \tilde{x}_2, \dots, \tilde{x}_{2B}$, and compute $z_i = f_{proj}(f_\theta(\tilde{x}_j))$ be the projected features of each augmented image $\Tilde{x_i}$. Let $y_i$ be the temporal class label of $\tilde{x}_i$ and $P(i) = \{p\in \{1, 2, \dots, 2B\} \backslash \{i\}: y_p = y_i\}$.
\begin{equation}
    \mathcal{L}_{TCL} = \sum_{i}\frac{-1}{|P(i)|} \sum_{p\in P(i)} \log \frac{\exp (z_i \cdot z_p / \tau)}{\sum_{a\neq i} \exp (z_i \cdot z_a / \tau)}.
\label{equation:supcon}
\end{equation}
\looseness=-10000
We refer to this as the temporal contrastive loss. It is conceptually similar to the temporal classification loss proposed in~\citep{orhan2020self}. However, in the temporal classification loss, the size of the classification layer needs to be gradually expanded as more data is processed by the model and more temporal classes are formed. Hence, the temporal contrastive loss is more flexible and more suitable for the streaming SSL setting.
\paragraph{Overall Loss Function.} In addition to the temporal contrastive loss, we also incorporate a standard self-supervised loss $\mathcal{L}_{SSL}$. In particular, we experimented with the SimCLR loss~\citep{chen2020simple, sohn2016improved} and the SimSiam loss~\citep{chen2021exploring} because they were shown to work well in lifelong self-supervised learning in prior works~\citep{zhuang2022well, purushwalkam2022challenges}.
The overall loss function is a sum of the temporal contrastive loss and the self-supervised contrastive loss $ \mathcal{L} = \mathcal{L}_{TCL} + \mathcal{L}_{SSL}$.
\paragraph{Warm-Start Training.} At the beginning of training, the model has only seen a very limited amount of data from the video stream. Even with a memory buffer, there is a likely high temporal correlation between the sampled frames which can cause instability in the training. To alleviate this problem, we warm-start the system by making no model updates on the first $M_{long}$ frames of the stream and just use them to fill the memory. The warm-start phase ensures that the model is trained on de-correlated samples from the buffer starting from the beginning.

\begin{minipage}[t]{0.45\textwidth}
\begin{algorithm}[H]
\caption{Temporal Segmentation}
\label{alg:temporalseg}
\definecolor{codeblue}{rgb}{0.25,0.5,0.5}
\lstset{
  backgroundcolor=\color{white},
  basicstyle=\fontsize{9pt}{9pt}\ttfamily\selectfont,
  columns=fullflexible,
  breaklines=true,
  captionpos=b,
  commentstyle=\fontsize{9pt}{9pt}\color{codeblue},
  keywordstyle=\fontsize{9pt}{9pt},
}
\begin{lstlisting}[language=python]
# n: number of clusters
# feats: features of the frames in the sequence
# F: maximization objective (defined by Equation 1). 
# Returns: detected change points in the stream (sorted)

def temporal_segment(n, feats, F):
    S = feats @ feats.T
    L = len(S)
    changepts = []
    for i in range(1, n):
        bestscore = 0
        for changept in range(1, L):
            temp = changepts + [changept]
            score = F(sorted(temp))
            if score > bestscore:
                bestscore = score
                bestchangept = changept
        changepts.append(bestchangept)
    return sorted(changepts)
\end{lstlisting}
\end{algorithm}
\end{minipage}
\hfill
\begin{minipage}[t]{0.52\textwidth}
\begin{algorithm}[H]
\caption{Memory Storyboard Streaming SSL}
\label{alg:memstoryboard}
\definecolor{codeblue}{rgb}{0.25,0.5,0.5}
\lstset{
  backgroundcolor=\color{white},
  basicstyle=\fontsize{9pt}{9pt}\ttfamily\selectfont,
  columns=fullflexible,
  breaklines=true,
  captionpos=b,
  commentstyle=\fontsize{9pt}{9pt}\color{codeblue},
  keywordstyle=\fontsize{9pt}{9pt},
}
\begin{lstlisting}[language=python]
# D: streaming data loader
# M_s: short-term memory buffer
# M_l: long-term memory buffer
# B_s, B_l: batch size for M_s, M_l
# T: default segment length
# r: subsampling rate

while True: # Loop until end of stream
    x = D.next()
    x_sub = subsample(x, r)
    M_l.add(x) # Updated with Reservoir
    M_s.add(x_sub) # Updated with FIFO
    if M_s[0].label > tc_label:
        tc_label = M_s[0].label
        n = len(M_s) / T
        feats = normalize(features(M_s))
        changes = temporal_segment(n, feats, F)
        update_labels(M_s, changes)
        update_labels(M_l, changes)
    data = sample(M_l, B_l, M_s, B_s)
    loss = TCL_loss(data) + SSL_loss(data)
    model.update(loss)
\end{lstlisting}
\end{algorithm}
\end{minipage}

\section{Experiments}

\subsection{Experiment Setup}
\paragraph{Datasets.}
We use two real-world egocentric video datasets in the experiments: (1) the child S subset of SAYCam dataset~\citep{sullivan2021saycam}, which contains 221 hours of video data collected from a head-mounted camera on the child from age 6-32 months, decoded at 25 fps;
(2) the KrishnaCam dataset~\citep{singh2016krishnacam}, which contains 70 hours of video data spanning nine months of the life of a graduate student, decoded at 10 fps. These two datasets have also been adopted in a number of existing self-supervised learning literature~\citep{orhan2020self, purushwalkam2022challenges, zhuang2022well, vong2024grounded}.

\paragraph{Training.} Following the architectural choices of Osiris~\citep{zhang2024integrating}, we use ResNet-50~\citep{he2016deep} as the feature extractor with group normalization~\citep{wu2018group} and the Mish activation function~\citep{misra2019mish}. Unless otherwise specified, the default hyperparameter values we use in our experiments are $b = 64$, $B=512$, $T=4.5K$ for SAYCam and $T=1.8K$ for KrishnaCam (both corresponding to 3 minutes of raw video), subsampling rate $r=8$ for SAYCam and $r=4$ for KrishnaCam. We train the models with two sets of memory sizes to evaluate their performance across different memory constraints: a larger memory constraint with $|M|=50K$, $|M_{short}|=5K$, $|M_{long}|=45K$, and a smaller memory constraint with $|M|=10K$, $|M_{short}|=1K$, $|M_{long}|=9K$. For context, there are a total of 18.2M frames in the SAYCam training set and 2.5M frames in the KrishnaCam training set. Therefore, even the large memory constraint of 50K frames only stores 0.27\% and 2.01\% of the total training frames in the memory buffer for SAYCam and KrishnaCam respectively. For the main experiments (Tables \ref{tab:main_table_saycam} and \ref{tab:main_table_kcam}), we employ label merging with $\tau = 0.998$ (i.e., the similarity threshold $\delta$ is the top $0.002$ quantile of the off-diagonal values in the similarity matrix); for all other experiments, label merging is not employed unless explicitly specified otherwise. Each experiment is run on one A100 GPU.

\paragraph{Evaluation.}
For object classification, we use \textit{mini}-ImageNet classification task for both SAYCam and KrishnaCam models. For each dataset, we also pick another downstream task that evaluates the learned representations of the training data itself. Evaluation tasks are summarized below.
\begin{itemize}
\item \textbf{\textit{mini}-ImageNet Classification:} Following a similar evaluation protocol as~\citet{zhuang2022well}, we evaluate the learned representations on downstream classification of a subsampled ImageNet~\citep{deng2009imagenet} dataset (\textit{mini}-INet). We extract the features of the model and train a support vector machine (SVM) to measure its classification performance. The \textit{mini}-ImageNet dataset contains 20K training images and 5K test images across 100 classes.
\item \textbf{ImageNet-1K and iNaturalist Classification:} Similar to the evaluation protocol used in~\citet{purushwalkam2022challenges}, we further evaluate the classification performance with a linear classifier on the larger ImageNet-1K~\citep{deng2009imagenet} dataset (INet) with 1.28M training images and 50K test images across 1K classes, and the iNaturalist-2018~\citep{van2018inaturalist} dataset (iNat) with 437K training images and 24K test images across 8142 classes.
\item \textbf{Labeled-S Classification:} For SAYCam models, we evaluate the classification performance on the Labeled-S dataset, following~\citet{orhan2020self}. The Labeled-S dataset is a labeled subset of the SAYCam frames, containing a total of 5786 images across 26 classes after 10x subsampling of frames. We randomly use 50\% as training data and 50\% as test data. 
\item \textbf{OAK Object Detection:} For KrishnaCam models, we evaluate the object detection performance on the Objects Around Krishna (OAK) dataset~\citep{wang2021wanderlust}, which includes bounding box annotations of 105 object categories on a subset of the KrishnaCam frames. We fine-tune the model on the entire training set of OAK for 10 epochs before evaluating on the OAK validation set, and report the AP50 metric.
\end{itemize}

\paragraph{Baselines.} We compare Memory Storyboard with a number of competitive SSL methods for image and video representation learning, and different memory buffer strategies:
\begin{itemize}
    \item \textbf{SimCLR:} In prior studies, \citet{zhuang2022well} showed that SimCLR~\citep{chen2020simple} is the strongest self-supervised learning method under streaming video setting, outperforming other SSL methods such as BYOL~\citep{grill2020bootstrap} and Barlow Twins~\citep{zbontar2021barlow}.
    \item \textbf{SimSiam:} In prior work, \citet{purushwalkam2022challenges} showed that SimSiam~\citep{chen2021exploring} is able to learn good representations from egocentric video data.
    \item \textbf{Osiris:} Osiris~\citep{zhang2024integrating} is a state-of-the-art unsupervised continual learning method that is developed towards static image sequences.
    \item \textbf{TC:} Temporal classification (TC)~\citep{orhan2020self} is a simple self-supervised learning method that is shown to work well on the SAYCam dataset under IID setting. It also uses temporal segments as a source of self-supervision; however, it does not actively group the frames together but instead relies on fixed intervals.
    \item \textbf{Reservoir Sampling:} We mainly use reservoir sampling~\citep{vitter1985random} as a default baseline approach for updating the memory buffer, which uniformly samples from all the seen images in the memory.
    \item \textbf{MinRed Buffer:} The minimum redundancy (MinRed) buffer~\citep{purushwalkam2022challenges} is a streaming self-supervised learning algorithm that alleviates temporal correlations in continuous video streams by storing minimally redundant samples in the replay buffer.
    \item \textbf{Two-tier Buffer:} As in MemStoryboard, we use a long-term memory updated with reservoir sampling and short-term memory updated with first-in-first-out (FIFO), but we do not apply the temporal contrastive loss or the temporal segmentation module.
\end{itemize}

\begin{table*}[t]
\centering
\begin{small}
\resizebox{0.99\linewidth}{!}{
\begin{tabular}{l|cccc|cccc}
\toprule
{\bf Method} &  {\bf \textit{mini}-INet} & {\bf INet} & {\bf iNat} & {\bf Labeled-S} & {\bf \textit{mini}-INet} & {\bf INet} & {\bf iNat} & {\bf Labeled-S} \\ 
\midrule
IID SimCLR~\citep{chen2020simple} & 44.04 & 30.44 & 8.69 & 59.50 & 44.04 & 30.44 & 8.69 & 59.50 \\
IID SimSiam~\citep{chen2021exploring} & 29.02 & 20.92 & 4.91 & 42.71 & 29.02 & 20.92 & 4.91 & 42.71 \\ \midrule
SimCLR No Replay & 5.76 & 2.22 & 0.07 & 19.13 & 5.76 & 2.22 & 0.07 & 19.13 \\ 
SimSiam No Replay & 6.44 & 1.47 & 0.04 & 22.03 & 6.44 & 1.47 & 0.04 & 22.03 \\ \midrule
\midrule
\multicolumn{1}{c}{} &
\multicolumn{4}{c}{\it Replay - 10k} & \multicolumn{4}{c}{\it Replay - 50k}\\ \midrule
Osiris~\citep{zhang2024integrating} & 31.16 & 19.48 & 4.68 & 45.81 & 36.90 & 23.16 & 5.85 & 50.88 \\
TC~\citep{orhan2020self} & 33.92 & 19.03 & 5.84 & 48.09 & 36.68 & 22.72 & 8.24 & 52.22 \\
SimCLR~\citep{chen2020simple} & 33.02 & 20.13 & 4.74 & 49.29 & 37.96 & 23.75 & 6.91 & 53.67 \\
\quad +MinRed~\citep{purushwalkam2022challenges} & 33.62 & 20.21 & 5.12 & 48.88 & 38.66 & 24.10 & 6.81 & 54.75 \\
\rowcolor{blue!10} \quad +Two-tier (Ours) & 33.80 & 20.70 & 5.61 & 49.05 & 39.22 & 24.93 & 7.07 & 55.43 \\
\rowcolor{blue!10} \quad +MemStoryboard (Ours) & 34.18 & 22.59 & 6.34 & \textbf{51.09} & 38.84 & 26.87 & 8.17 & \textbf{56.26} \\
SimSiam~\citep{chen2021exploring} & 20.90 & 13.72 & 2.55 & 39.12 & 26.66 & 14.44 & 3.79 & 43.09 \\
\quad +MinRed~\citep{purushwalkam2022challenges} & 22.68 & 17.85 & 3.17 & 39.78 & 25.58 & 18.99 & 4.24 & 40.37 \\
\rowcolor{blue!10} \quad +Two-tier (Ours) & 21.78 & 16.87 & 2.76 & 39.19 & 28.34 & 20.24 & 3.99 & 42.95 \\
\rowcolor{blue!10} \quad +MemStoryboard (Ours) & \textbf{36.86} & \textbf{26.70} & \textbf{8.46} & 49.87 & \textbf{41.46} & \textbf{28.92} & \textbf{10.41} & 53.78 \\
\bottomrule
\end{tabular}
}
\end{small}
\caption{\textbf{Results on streaming SSL from SAYCam~\citep{sullivan2021saycam}.} Downstream evaluation on object classification (Accuracy \%) for SSL models trained under the streaming setting. For ``No Replay'' and ``IID'' the results are the same for different memory buffer sizes. The ``IID'' methods are not under the streaming setting and are for reference only as a performance ``upper bound'' with the same number of gradient updates. Unless specified, standard reservoir sampling is used in the replay buffer.}
\label{tab:main_table_saycam}
\end{table*}

\begin{table*}[t]
\centering
\begin{small}
\begin{tabular}{l|cccc|cccc}
\toprule
{\bf Method} & {\bf \textit{mini}-INet} & {\bf INet} & {\bf iNat} & {\bf OAK} & {\bf \textit{mini}-INet} & {\bf INet} & {\bf iNat} & {\bf OAK} \\ \midrule
IID SimCLR~\citep{chen2020simple} & 36.90 & 23.77 & 5.60 & 39.54 & 36.90 & 23.77 & 5.60 & 39.54 \\
IID SimSiam~\citep{chen2021exploring} & 28.58 & 22.28 & 4.16 & 44.86 & 28.58 & 22.28 & 4.16 & 44.86 \\ \midrule
SimCLR No Replay & 4.84 & 1.35 & 0.07 & 14.01 & 4.84 & 1.35 & 0.07 & 14.01 \\
SimSiam No Replay & 8.88 & 1.92 & 0.05 & 27.34 & 8.88 & 1.92 & 0.05 & 27.34 \\ \midrule
\midrule
\multicolumn{1}{c}{} &
\multicolumn{4}{c}{\it Replay - 10k} & \multicolumn{4}{c}{\it Replay - 50k}\\ \midrule
Osiris~\citep{zhang2024integrating} & 30.10 & 19.03 & 3.55 & 32.25 & 32.38 & 20.85 & 3.96 & 33.78 \\
TC~\citep{orhan2020self} & 32.58 & 19.19 & 6.01 & 32.61 & 32.94 & 20.50 & 6.25 & 28.56 \\
SimCLR~\citep{chen2020simple} & 31.46 & 19.09 & 4.43 & 31.92 & 34.98 & 22.37 & 5.19 & 33.30 \\
\quad +MinRed~\citep{purushwalkam2022challenges} & 31.56 & 19.93 & 4.69 & 34.78 & 34.84 & 22.29 & 5.30 & 35.65 \\
\rowcolor{blue!10} \quad +Two-tier (Ours) & 33.26 & 20.39 & 5.04 & 33.72 & 35.78 & 22.42 & 5.29 & 35.68 \\
\rowcolor{blue!10} \quad +MemStoryboard (Ours) & \textbf{33.30} & 22.52 & 5.65 & 36.01 & \textbf{36.08} & 25.37 & 6.28 & 38.58 \\
SimSiam~\citep{chen2021exploring} & 19.16 & 12.94 & 2.85 & 39.38 & 21.84 & 14.13 & 3.56 & 41.13 \\
\quad +MinRed~\citep{purushwalkam2022challenges} & 20.90 & 14.53 & 3.16 & 43.74 & 22.88 & 17.64 & 5.12 & 44.17 \\
\rowcolor{blue!10} \quad +Two-tier (Ours) & 20.08 & 13.76 & 2.91 & 43.68 & 22.14 & 17.06 & 3.73 & 44.41 \\
\rowcolor{blue!10} \quad +MemStoryboard (Ours) & 33.22 & \textbf{23.76} & \textbf{6.52} & \textbf{45.18} & 34.62 & \textbf{25.52} & \textbf{6.78} & \textbf{46.17} \\
\bottomrule
\end{tabular}
\end{small}
\caption{\textbf{Results on streaming SSL from KrishnaCam~\citep{singh2016krishnacam}.} Downstream evaluation on object classification (Accuracy \%) and object detection (AP50 \%) for SSL models trained under the streaming setting. The structure of the table is otherwise similar to Table~\ref{tab:main_table_saycam}.
}
\label{tab:main_table_kcam}
\end{table*}

\subsection{Main Results}
\looseness=-10000
In Tables~\ref{tab:main_table_saycam} and~\ref{tab:main_table_kcam}, we report the main results on streaming SSL on both SAYCam and KrishnaCam. Firstly, we observe that all SSL methods work poorly in the streaming setting without replay, and larger memory leads to better performance. In terms of memory buffer strategies, our two-tier memory hierarchy and MinRed~\citep{purushwalkam2022challenges} outperform reservoir sampling. 

Memory Storyboard achieves superior performance in all readout tasks compared to other streaming SSL models. For SimCLR-based methods, Memory Storyboard considerably narrows the gap between streaming learning and IID training. Memory Storyboard also significantly outperforms all baseline methods with a considerable gap on both the ImageNet classification and the challenging OAK object detection benchmark. For SimSiam-based methods, Memory Storyboard not only outperforms all streaming learning baselines by a considerable margin but also beats IID SimSiam training on all readout tasks. 

Memory Storyboard with SimSiam achieves the overall best performance across different training datasets and evaluation metrics. We hypothesize that Memory Storyboard works better with SimSiam~\citep{chen2021exploring} than SimCLR~\citep{chen2020simple} in our experiments because the SimCLR loss can be conflicting with the temporal contrastive objective. SimCLR treats some highly correlated images in the same batch as negative samples during training, despite the fact that the other images in the batch sampled from the short-term buffer might be very similar to the current image since they are temporally close to each other. This issue is exacerbated in the SAYCam experiments due to the high frequency (25 fps) of the SAYCam video stream. By incorporating the temporal contrastive loss in Memory Storyboard, we successfully address this issue by utilizing only images in other temporal classes as negative samples.

\begin{figure*}[t]
    \centering
    \includegraphics[width=0.95\linewidth,trim={0.5cm 0 0 0},clip]{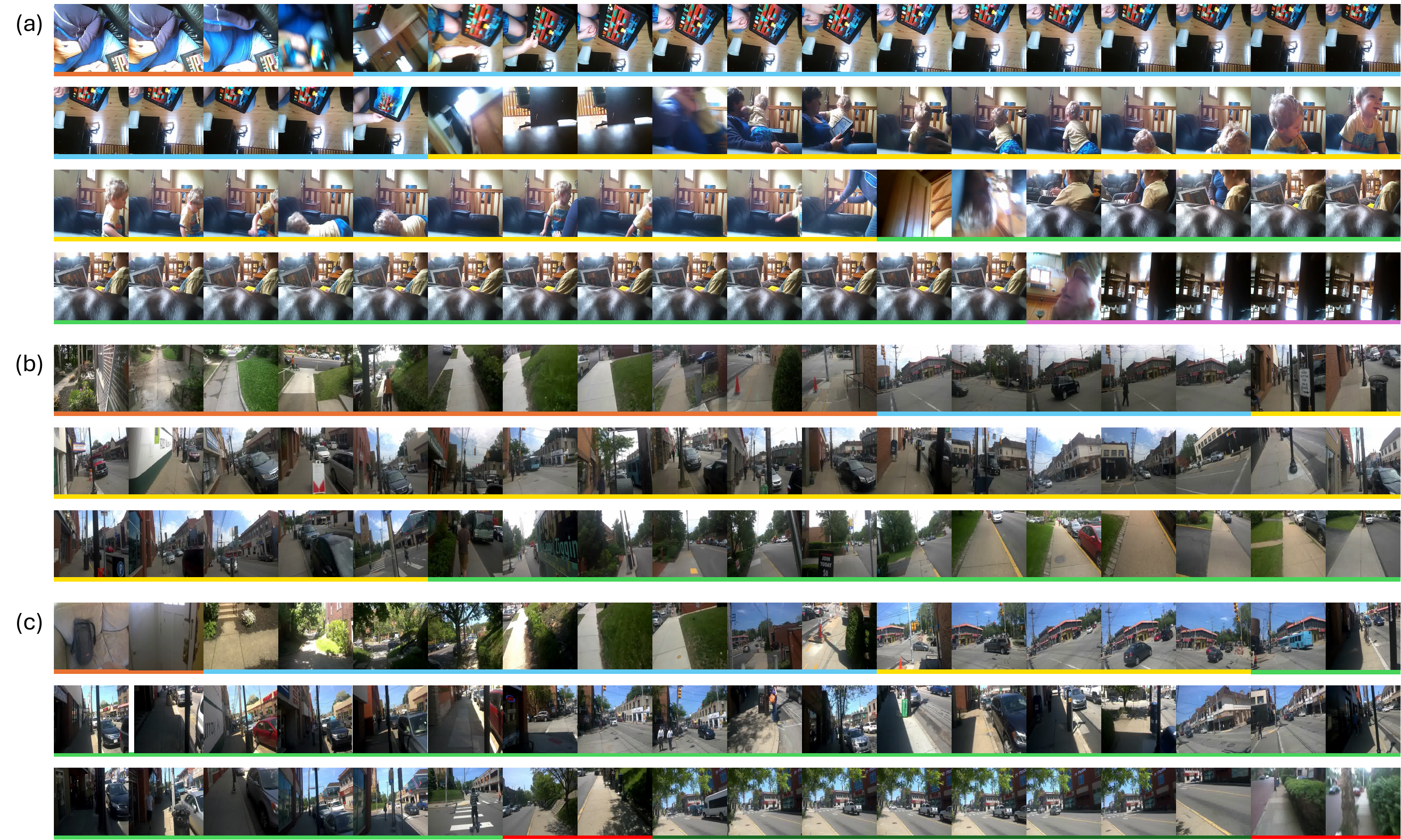}
    \caption{
    \textbf{Visualization of the temporal segments produced by Memory Storyboard on (a) SAYCam (b)(c) KrishnaCam at the end of training.} The images are sampled at 10 seconds per frame. Each color bar corresponds to a temporal class (the first and the last class might be incomplete). Temporal segments produced at the beginning of training are provided in the appendix for comparison.}
    \label{fig:saycam_seg_vis}
\end{figure*}

Overall, the results demonstrate that Memory Storyboard is effective at learning good representations from a streaming video source, and the learned representations can be successfully transferred to downstream vision tasks on the training dataset itself or an external dataset.
\paragraph{Qualitative Results.} We visualize the temporal segments produced by Memory Storyboard at the end of training in Figure~\ref{fig:saycam_seg_vis}. The results demonstrate that our temporal segmentation module can produce semantically meaningful temporal segments, showing its strong temporal abstraction capability. We emphasize that the representations are entirely developed during the streaming SSL training as the networks are trained from scratch. We provide additional qualitative results in Appendix~\ref{sec:supp_exp_results}.

\subsection{Ablation Experiments and Other Training Factors}

\looseness=-10000
In this section, we study how varying different training factors affect the performance of Memory Storyboard, including label merging, subsampling rate, and average segment length. We use SimCLR~\citep{chen2020simple} as the base SSL method for training, and a long-term memory size of 50K unless otherwise specified. Please refer to Appendix~\ref{sec:supp_exp_results} for additional ablation experiments.

\begin{table*}[t]
\centering
\begin{small}
\resizebox{0.99\linewidth}{!}{
\begin{tabular}{l|c|cccc|cccc}
\toprule
{\bf Base SSL Method} & {\bf Label Merging} & {\bf \textit{mini}-INet} & {\bf INet} & {\bf iNat} & {\bf Labeled-S} & {\bf \textit{mini}-INet} & {\bf INet} & {\bf iNat} & {\bf Labeled-S} \\ 
\midrule
\multicolumn{1}{c}{} &\multicolumn{1}{c}{} &
\multicolumn{4}{c}{\it Replay - 10k} & \multicolumn{4}{c}{\it Replay - 50k}\\ \midrule
SimCLR & \xmark & 35.02 & 20.72 & 5.65 & \textbf{51.33} & 39.58 & 24.78 & 7.77 & \textbf{56.29} \\
SimCLR & \cmark & 34.18 & 22.59 & 6.34 & 51.09 & 38.84 & 26.87 & 8.17 & 56.26 \\ \midrule
SimSiam & \xmark & 36.72 & 22.99 & 6.66 & 49.12 & 41.32 & 26.37 & 9.85 & 53.29 \\
SimSiam & \cmark & \textbf{36.86} & \textbf{26.70} & \textbf{8.46} & 49.87 & \textbf{41.46} & \textbf{28.92} & \textbf{10.41} & 53.78 \\
\bottomrule
\end{tabular}
}
\end{small}
\caption{Ablation on label merging for MemStoryboard trained on SAYCam.}
\label{tab:ablation_saycam_merging}
\end{table*}

\begin{table*}[t]
\centering
\begin{small}
\resizebox{0.95\linewidth}{!}{
\begin{tabular}{l|c|cccc|cccc}
\toprule
{\bf Base SSL Method} & {\bf Label Merging} & {\bf \textit{mini}-INet} & {\bf INet} & {\bf iNat} & {\bf OAK} & {\bf \textit{mini}-INet} & {\bf INet} & {\bf iNat} & {\bf OAK} \\ \midrule
\multicolumn{1}{c}{} & \multicolumn{1}{c}{} &
\multicolumn{4}{c}{\it Replay - 10k} & \multicolumn{4}{c}{\it Replay - 50k}\\ \midrule
SimCLR & \xmark & 33.72 & 20.13 & 5.64 & 35.77 & \textbf{36.36} & 22.75 & 6.10 & 38.67 \\
SimCLR & \cmark & 33.30 & 22.52 & 5.65 & 36.01 & 36.08 & 25.37 & 6.28 & 38.58 \\ \midrule
SimSiam & \xmark & 33.78 & 21.38 & 6.51 & \textbf{45.33} & 35.20 & 22.75 & 6.71 & \textbf{46.64} \\
SimSiam & \cmark & \textbf{33.22} & \textbf{23.76} & \textbf{6.52} & 45.18 & 34.62 & \textbf{25.52} & \textbf{6.78} & 46.17 \\
\bottomrule
\end{tabular}
}
\end{small}
\caption{Ablation on label merging for MemStoryboard trained on KrishnaCam.}
\label{tab:ablation_kcam_merging}
\end{table*}

\paragraph{Label Merging.}
We train MemStoryboard with and without the label merging mechanism and present the results in Tables~\ref{tab:ablation_saycam_merging} and~\ref{tab:ablation_kcam_merging}. We observe that incorporating label merging consistently improves performance on the ImageNet and iNaturalist classification tasks, which are out-of-domain classification settings. This suggests that grouping labels for representation learning is helpful in recognizing new classes at test time. While the improvements are generally modest in absolute terms, they are robust across different datasets and buffer sizes. Performance on the other benchmarks (\textit{mini}-ImageNet, Labeled-S, and OAK) remains largely stable with or without label merging.

\paragraph{Subsampling Rate.} We train Memory Storyboard with different subsampling rates when adding data fetched from the current stream to the short-term memory. Results are shown in Table~\ref{tab:subsample}. A subsampling ratio of 8 works best for SAYCam, while a ratio of 4 works best for KrishnaCam. Since the two datasets are decoded at different frequencies (25 fps for SAYCam and 10 fps for KrishnaCam), the effective frequency of frames entering the short-term buffer is 3.13 and 2.50 fps respectively. The result suggests that an effective frequency of around 3 fps is preferable although the optimal subsample ratio is dependent on the nature of the video stream. Intuitively, when the subsampling ratio is too small, the images entering the short-term buffer may have too much temporal correlation and hence would hurt the performance; when the subsampling ratio is too big, the model skips too many frames without training on them and the temporal clustering may also become less precise.

\begin{minipage}[t]{0.48\textwidth}
\begin{table}[H]
\centering
\resizebox{0.99\textwidth}{!}{
\begin{tabular}{l|cc|cc}
\toprule
{\bf Subsample} & \multicolumn{2}{c|}{\bf SAYCam} & \multicolumn{2}{c}{\bf KrishnaCam} \\ 
{\bf Ratio} & \textit{mini}-INet & Labeled-S & \textit{mini}-INet & OAK AP50 \\ \midrule
1$\times$ & 36.70 & 55.29 & 35.54 & 38.55 \\
2$\times$ & 37.18 & 55.43 & 35.60 & 37.38 \\
4$\times$ & 38.38 & 55.84 & \textbf{36.36} & 38.67 \\
8$\times$ & \textbf{39.58} & \textbf{56.29} & 35.48 & \textbf{38.90} \\
16$\times$ & 38.62 & 55.81 & 35.88 & 38.22 \\
\bottomrule
\end{tabular}
}
\vspace{-0.05in}
\caption{Effect of subsampling ratio for $M_{short}$ in Memory Storyboard.}
\label{tab:subsample}
\end{table}
\end{minipage}
\hfill
\begin{minipage}[t]{0.48\textwidth}
\begin{table}[H]
\centering
\resizebox{0.92\textwidth}{!}{
\begin{tabular}{l|cc|cc}
\toprule
\multirow{2}{*}{\bf $T$} & \multicolumn{2}{c|}{\bf SAYCam} & \multicolumn{2}{c}{\bf KrishnaCam} \\ 
 & \textit{mini}-INet & Labeled-S & \textit{mini}-INet & OAK AP50 \\ \midrule
1 min & 38.90 & 55.05 & 35.86 & 38.57 \\
2 min & 39.16 & 56.53 & 36.30 & \textbf{38.68} \\
3 min & \textbf{39.58} & 56.29 & \textbf{36.36} & 38.67 \\
5 min & 39.26 & \textbf{56.64} & 36.28 & 38.07 \\
10 min & 38.34 & 55.36 & 35.98 & 37.53 \\
\bottomrule
\end{tabular}
}
\caption{Performance of Memory Storyboard using different average temporal segment lengths.}
\label{tab:seg_length}
\end{table}
\end{minipage}
\paragraph{Average Segment Length.} We trained Memory Storyboard with different average segment lengths $T$ ranging from 1 minute to 10 minutes on SAYCam and KrishnaCam. The results are shown in Table~\ref{tab:seg_length}. We demonstrate that the performance of Memory Storyboard is generally robust to average segment length (which determines the number of temporal segments in the segmentation module). We also find that the performance on downstream tasks becomes worse when the average segment length is very long $(T=10\text{ min})$ on both datasets. This observation is different from that of temporal classification~\citep{orhan2020self} which claims longer segments are more helpful.

\begin{figure*}[t]
\vspace{-0.3in}
\centering
\includegraphics[width=0.99\textwidth]{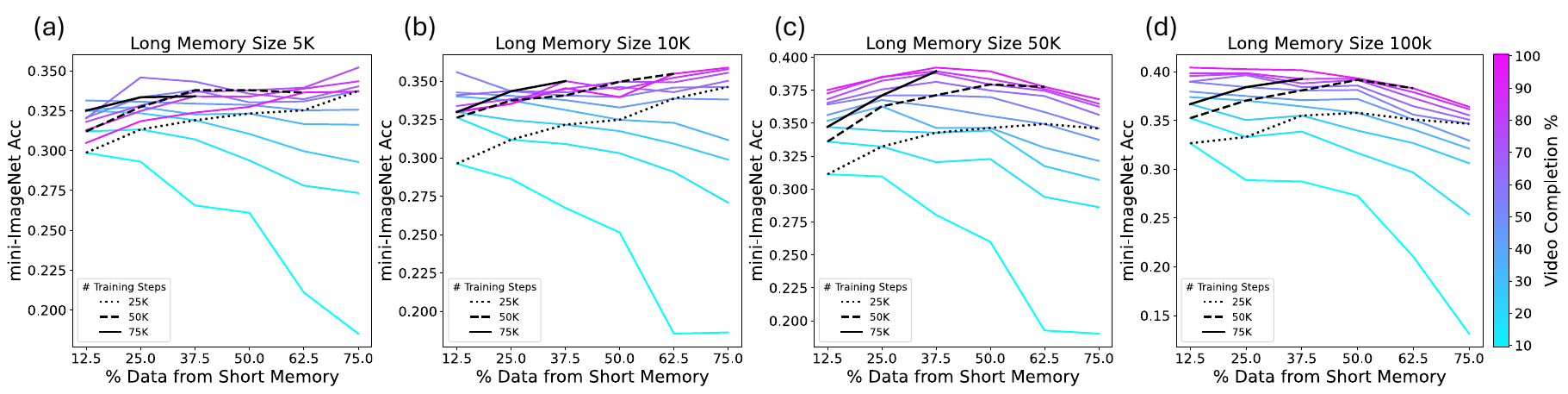}
    \caption{
    \textbf{Memory Storyboard model performance on SAYCam with different long-term memory sizes} (5k, 10k, 50k, and 100k) \textbf{and varying training batch compositions} (12.5\% -- 75.0\% from $M_{short}$) using SVM readout. Each colored line represents the performance of different training batch compositions when \textbf{the model has seen the same amount of data} from the stream. Each black line represents the performance of different training batch compositions when the model has taken \textbf{the same number of gradient updates}.}
    \label{fig:batch_buffer}
\end{figure*}

\subsection{Optimal Batch Composition Under Different Memory Constraints}
\label{sec:batch_buffer}
\looseness=-10000
In Memory Storyboard, the training batch is composed of samples from both the long-term memory and the short-term memory (see Figure~\ref{fig:detail}). However, the optimal composition ratio of the training batch, i.e. the optimal percentage of data in the training batch that comes from the short-term memory, is yet to be explored. Sampling more data from the short-term memory means we can digest more data within a fixed number of training steps, but there will be more distribution shift between different training batches. On the other hand, sampling more data from the long-term memory buffer may result in overfitting on the long-term memory data. In this section, we experiment with different memory sizes and training composition and demonstrate the optimal batch composition under different memory constraints.

\looseness=-10000
We fix the size of the short-term memory $|M_{short}|$ to be $5K$ and vary the memory constraint for the long-term memory $|M_{long}|=5K, 10K, 50K, 100K$. For each long-term memory size, we experiment with batch size from data stream $b=64, 128, 192, 256, 320, 384$ (which corresponds to 12.5\% though 75\% of the training batch size). We sample $b$ images from the short-term memory and $512-b$ images from the long-term memory to compose a training batch. We evaluate the model with SVM readout on \textit{mini-}ImageNet after the model has seen every 10\% of the entire data stream and plot the results in Figure~\ref{fig:batch_buffer}. We discuss the different observations for large memory size and small memory size respectively.
\begin{itemize}
    \item \textbf{Large $|M_{long}|$:} When $|M_{long}|$ is large (Figures~\ref{fig:batch_buffer}(c) and~\ref{fig:batch_buffer}(d)), overfitting on the memory is unlikely and hence we can sample more data from the long-term memory and the performance still keeps increasing as the model sees more data. Hence, with the same amount of data seen by the model (colored curves), it is better to sample only a small batch from the short-term memory. However, when we control the number of model update steps to the same (black curves), neither focusing on the short-term memory nor focusing on the long-term memory is preferable. In such cases, the optimal batch size from the short-term memory is at roughly 50\% of the training batch.
    \looseness=-10000 
    \item \textbf{Small $|M_{long}|$:} When $|M_{long}|$ is small (Figures~\ref{fig:batch_buffer}(a) and~\ref{fig:batch_buffer}(b)), the model is prone to overfitting on the memory. As a result, with the same number of model update steps (black curves), taking more images from $M_{short}$ gives better results. With the same amount of data seen by the model (colored curves), getting a higher percentage of data from $M_{long}$ has an advantage in the beginning when there is less memory overfitting. Ultimately, focusing on $M_{short}$ is more beneficial in the late stage.
\end{itemize}
To summarize, the optimal training batch composition depends on memory and compute constraints. More samples from the long-term memory are preferred when a large memory is available (e.g., 50K images from a 200-hour stream) and model performance is evaluated after seeing a fixed amount of data. More samples from the short-term memory are preferred when memory is limited to prevent overfitting on the long-term memory data. When both memory and compute are sufficient and performance is measured under a fixed compute budget, a balanced batch composition is most effective for real-time learning.

\section{Conclusion}

The ability to continuously learn from large-scale uncurated streaming video data is crucial for applying self-supervised learning methods in real-world embodied agents. Existing works have limited exploration of this problem, have mainly focused on static datasets, and do not perform well in the streaming video setting. Inspired by the event segmentation mechanism in human cognition, in this work, we propose Memory Storyboard, which leverages temporal segmentation to produce a two-tier memory hierarchy akin to the short-term and long-term memory of humans. Memory Storyboard combines a temporal contrastive objective and a standard self-supervised contrastive objective to facilitate representation learning from scratch through streaming video experiences. Memory Storyboard achieves state-of-the-art performance on downstream classification and object detection tasks when trained on real-world large egocentric video datasets. By studying the effects of subsampling rates, average segment length, normalization, and optimal batch composition under different compute and memory constraints, we also offer valuable insights on the design choices for streaming self-supervised learning.

\section*{Acknowledgements}
We thank James McClelland for a helpful discussion on event segmentation and pointers to relevant literature. We thank Peiqi Liu and Anh Ta for their explorations on training self-supervised learning models on the SAYCam dataset. The work is supported in part by the Institute of Information \& Communications Technology Planning \& Evaluation (IITP) under grant RS-2024-00469482, funded by the Ministry of Science and ICT (MSIT) of the Republic of Korea in connection with the Global AI Frontier Lab International Collaborative Research. YY is supported by the Meta AI Mentorship Program. The compute is supported through the NYU IT High Performance Computing resources, services, and staff expertise.

\bibliography{collas2025_conference}
\bibliographystyle{collas2025_conference}

\clearpage
\appendix
\section{Experiment Details}
\label{sec:supp_exp_details}
\paragraph{Model Architecture.} On top of the ResNet backbone, we use a two-layer MLP with 2048 hidden units, 128 output units, and ReLU activation function as the projector. In Memory Storyboard, we create two separate projectors for $\mathcal{L}_{TCL}$ and $ \mathcal{L}_{SSL}$.

\paragraph{Training.}
\looseness=-10000
For all experiments in Tables~\ref{tab:main_table_saycam} and~\ref{tab:main_table_kcam}, we used a total batch size of 512 (64 from $M_{short}$ and 448 from $M_{long}$ by default). The input resolution of the images to the model is 112. We apply a standard data augmentation pipeline for SSL methods following~\citet{zhuang2022well}, which include random resized crop, random horizontal flip, random color jitter, random grayscale, random Gaussian filter, and color-normalization with ImageNet~\citep{deng2009imagenet}. For the SimCLR~\citep{chen2020simple}, Osiris~\citep{zhang2024integrating}, and TC~\citep{orhan2020self} experiments, we used the Adam~\citep{kingma2014adam} optimizer with a constant learning rate of 0.001, and a projector with 2 MLP layers of size 2048 and 128 respectively. For the SimSiam~\citep{chen2021exploring} experiments, we used the SGD optimizer with learning rate 0.05, momentum 0.9, and weight decay 1e-4, and a projector with 3 MLP layers of size 2048.

\paragraph{Evaluation.} For \textit{mini}-ImageNet and Labeled-S evaluations, the streaming SSL models are evaluated every 5\% of the entire dataset. That is, we store 20 model checkpoints throughout the streaming training and evaluate them on \textit{mini}-ImageNet and Labeled-S with SVM readout. The \textit{best} results among these checkpoints are reported. Similar to~\citet{zhuang2022well}, for SVM readout, we report the best performance among learning rate values \{1e-7, 1e-6, 1e-5, 1e-4, 1e-3, 1e-2, 1e-1, 1, 1e1, 1e2\}.

For ImageNet-1K and iNaturalist-2018 evaluations, we evaluate the final model after streaming SSL training on the entire dataset. Following~\citet{purushwalkam2022challenges}, we train a linear classifier on top of the normalized learned representations and report the classification accuracy. We used a batch size of 1024. For ImageNet-1K, we used the LARS~\citep{you2017large} optimizer with learning rate 3.0, momentum 0.9, and cosine learning rate schedule for 10 epochs. For iNaturalist-2018, we used the LARS~\citep{you2017large} optimizer with learning rate 12.0, momentum 0.9, and cosine learning rate schedule for 20 epochs.

For OAK evaluations, we use Faster R-CNN~\citep{ren2015faster}, a popular two-stage object detector. We initialize the ResNet-50~\citep{he2016deep} backbone with the backbone of the final checkpoint of the streaming SSL model, and fine-tune the entire model on OAK with IID training for 10 epochs, following the training configurations of~\citep{wu2023label}.

\section{Additional Related Work}
\label{sec:supp_related}
\paragraph{Self-Supervised Learning.}
\looseness=-10000
A large number of self-supervised representation learning methods in computer vision follows the contrastive learning framework~\citep{oord2018representation, misra2020self, tian2020contrastive, he2020momentum, chen2020simple, chen2021exploring} which maximizes the agreement of representations of two augmented views of the same image and minimizes that of different images. Extending this idea, the supervised contrastive (SupCon) method~\citep{khosla2020supervised} uses the labels as an extra supervision signal to get multiple positive crops for each anchor image. Other recent self-supervised learning works include pretext tasks~\citep{doersch2015unsupervised, noroozi2016unsupervised, gidaris2018unsupervised, pathak2016context}, feature space clustering~\citep{caron2018deep, caron2020unsupervised, ren2021online}, distillation with asymmetric architectures~\citep{grill2020bootstrap, chen2021exploring}, redundancy reduction~\citep{zbontar2021barlow, bardes2021vicreg}, and masked autoencoding~\citep{he2022masked}. Most relevant of these to our work,~\citet{orhan2020self} proposes the temporal classification objective, which outperforms contrastive learning objectives on the SAYCam dataset~\citep{sullivan2021saycam}. Our work enhances the temporal classification method by using a more flexible supervised contrastive objective, and leveraging temporal segmentation~\citep{potapov2014category, afham2023revisiting}, which have been used extensively in video summarization~\citep{zhu2020dsnet, zhang2016video, rochan2018video}.

\paragraph{Temporal Segmentation in Human Cognition.} Prior research in psychology and cognitive sciences has shown that humans, including infants, are able to identify boundaries between action segments~\citep{newtson1977objective, zacks2001perceiving, baldwin2001infants, saylor2007infants, baldassano2017discovering, yates2022neural}. Evidence in neuro-imaging further shows that event segmentation is an automatic component in human perception~\citep{zacks2007event}. Temporal event segmentation has proven to be critical for memory formation and retrieval~\citep{lassiter1991unitization, ezzyat2011constitutes, dubrow2013influence, silva2019rapid, sasmita2022measuring}. The temporal segmentation component in our proposed framework is motivated by how humans interpret videos as segments with coherent semantics. We demonstrate that temporal segmentation can improve the learned visual representation.
\section{Additional Results}
\label{sec:supp_exp_results}

\subsection{Performance of Different Normalization Layers}
We experimented with a variation of Memory Storyboard as well as three baseline methods (SimCLR~\citep{chen2020simple}, Osiris~\citep{zhang2024integrating}, and Temporal Classification~\citep{orhan2020self}) where the group normalization layers in the ResNet backbone are replaced with batch normalization~\citep{ioffe2015batch} layers. The models are trained on SAYCam and evaluated on the downstream \textit{mini}-ImageNet classification task with an SVM. The resulting accuracies are shown in Table~\ref{tab:normalization}. We observe that GroupNorm significantly outperforms BatchNorm for all the models examined. This result is aligned with the conclusion in~\citep{zhang2024integrating} that BatchNorm is less compatible with unsupervised continual learning, and extends the conclusion to streaming SSL.

\begin{table}[h]
\centering
\begin{tabular}{ccccc}
\toprule
 & SimCLR & Osiris & TC & MemStoryboard \\
\midrule
Batch Norm & 33.62 & 33.32 & 33.16 & 33.68 \\
Group Norm & \textbf{37.96} & \textbf{36.90} & \textbf{36.68} & \textbf{39.58} \\
\bottomrule
\end{tabular}
\caption{
Group norm is better at dealing with temporal non-stationarity for streaming SSL.}
\label{tab:normalization}
\end{table}

\subsection{Separating Short-term Memory Batch and Long-term Memory Batch}
Inspired by the design of separating the loss on the new data and the replay data in Osiris~\citep{zhang2024integrating}, we investigate the optimal strategy of applying the temporal contrastive loss on the training batch. We consider applying the temporal contrastive loss only on data from short-term memory, only on data from long-term memory, separately on data from short-term and long-term memory and average the losses, and on the entire training batch (concatenated data from short-term and long-term memory). We report the results in Table~\ref{tab:sep_losses}. For experiments in the main paper, we apply the temporal contrastive loss only on data from long-term memory.

The results here demonstrate that applying the temporal contrastive loss only on data from long-term memory or on the entire training batch achieves the best performance. Applying the temporal contrastive loss only on data from short-term memory achieves inferior performance due to the limited number of temporal classes in the short-term buffer.

\begin{table}[H]
\centering
\begin{tabular}{l|cc|cc}
\toprule
 & \multicolumn{2}{c|}{\bf SAYCam} & \multicolumn{2}{c}{\bf KrishnaCam} \\ 
 & \textit{mini}-ImageNet & Labeled-S & \textit{mini}-ImageNet & OAK mAP \\ \midrule
Short Only & 38.54 & 52.95 & 34.98 & 19.53 \\
Long Only & \textbf{39.58} & \textbf{56.29} & 36.36 & 21.29 \\
Concatenate & 38.34 & 55.43 & 36.08 & 21.20 \\
Separate & 39.42 & 54.95 & \textbf{36.70} & \textbf{21.40} \\
\bottomrule
\end{tabular}
\caption{Performance of Memory Storyboard when the temporal contrastive loss is applied on different parts of the training batch.}
\label{tab:sep_losses}
\end{table}

\subsection{Memory Storyboard in the IID Setting}
To provide more context for the performance of Memory Storyboard in the Streaming Learning setting, we investigate the performance of Memory Storyboard in the IID setting. We take a SimCLR IID pre-trained model and use it to generate temporal segmentations on the entire dataset. Then we assign pseudo-labels to each frame according to the temporal segmentation results and train Memory Storyboard in the IID setting, taking the same number of gradient steps as the streaming setting. We observe that IID Memory Storyboard outperforms IID SimCLR and IID Simsiam, demonstrating the effectiveness of the temporal contrastive loss even in the IID setting.
\begin{table}[H]
\centering
\small
\begin{tabular}{l|ccc|ccc}
\toprule
\multirow{2}{*}{\bf Method} & \multicolumn{3}{c|}{\bf SAYCam} & \multicolumn{3}{c}{\bf KrishnaCam} \\ 
& \textit{mini}-INet & INet & iNat & \textit{mini}-INet & INet & iNat \\ \midrule
SimCLR MemStoryboard (50K) & 39.58 & 24.78 & 7.77 & 36.36 & 22.75 & 6.10 \\
SimCLR MemStoryboard (50K) & 41.32 & 26.37 & 9.85 & 35.20 & 22.75 & 6.71 \\
IID SimCLR & 44.04 & 30.44 & 8.69 & 36.90 & 23.77 & 5.60 \\
IID SimSiam & 29.02 & 20.92 & 4.91 & 28.58 & 22.28 & 4.16 \\
IID SimCLR MemStoryboard & 44.54 & 29.98 & 8.92 & 37.60 & 24.43 & 7.36 \\
IID SimSiam MemStoryboard & 42.30 & 31.02 & 10.07 & 36.54 & 26.60 & 7.06 \\
\bottomrule
\end{tabular}
\caption{Performance of Memory Storyboard in the IID Setting.}
\label{tab:storyboard_iid}
\end{table}

\subsection{Memory Storyboard with the Cross Entropy Loss}
One alternative objective for the temporal contrastive loss in Memory Storyboard is the cross entropy (CE) loss.
We investigate the performance of Memory Storyboard with the CE loss instead of the Supervised Contrastive (SupCon) loss. Results are summarized in Table~\ref{tab:ce_loss}. We observe that the CE loss outperforms the SupCon loss as the temporal contrastive loss when trained jointly with the SimCLR objective. We opted for the SupCon loss in the main text due to its flexibility. With the CE loss, we either need to know the number of temporal segments beforehand or gradually increase the size of the final classifier layer as the model sees more data, which is not ideal for streaming learning on a never-ending video stream. With the SupCon loss, we can learn from a very large number of temporal segments with the same model size.

\begin{table}[H]
\centering
\resizebox{0.99\textwidth}{!}{
\begin{tabular}{l|ccc|ccc|ccc|ccc}
\toprule
\multirow{2}{*}{\bf Temp. Contrast Loss} & \multicolumn{3}{c|}{\bf SAYCam 10K} & \multicolumn{3}{c|}{\bf SAYCam 50K} & \multicolumn{3}{c|}{\bf KrishnaCam 10K} & \multicolumn{3}{c}{\bf KrishnaCam 50K} \\ 
& \textit{mini}-INet & INet & iNat & \textit{mini}-INet & INet & iNat & \textit{mini}-INet & INet & iNat & \textit{mini}-INet & INet & iNat \\ \midrule
SupCon & 35.02 & 20.72 & 5.65 & 39.58 & 24.78 & 7.77 & 33.72 & 20.13 & 5.64 & 36.36 & 22.75 & 6.10 \\
Cross Entropy & 35.58 & 24.04 & 6.62 & 40.06 & 26.92 & 8.19 & 34.44 & 25.27 & 5.95 & 36.26 & 26.67 & 6.73 \\
\bottomrule
\end{tabular}
}
\caption{Results on Memory Storyboard (using SimCLR as the base SSL method) with the cross entropy-loss instead of the supervised contrastive loss as the temporal contrastive loss.}
\label{tab:ce_loss}
\end{table}

\subsection{Memory Storyboard with only the Temporal Contrastive Loss}
To demonstrate the need for both the self-supervised loss and the temporal contrastive loss in our training objective, we experiment with using only 
the temporal contrastive (SupCon) loss and not the self-supervised loss (only using the self-supervised loss (SimCLR or SimSiam) and not the temporal contrastive loss has been experimented in the “two-tier” buffer baseline in Tables~\ref{tab:main_table_saycam}~and~~\ref{tab:main_table_kcam}). Results are shown in Table~\ref{tab:supcon_only}. We observe that the performance of only using the SupCon loss is also inferior to the full memory storyboard method, demonstrating the necessity of joint training on both losses for best performance.
\begin{table}[H]
\centering
\resizebox{0.99\textwidth}{!}{
\begin{tabular}{l|ccc|ccc|ccc|ccc}
\toprule
\multirow{2}{*}{\bf Method} & \multicolumn{3}{c|}{\bf SAYCam 10K} & \multicolumn{3}{c|}{\bf SAYCam 50K} & \multicolumn{3}{c|}{\bf KrishnaCam 10K} & \multicolumn{3}{c}{\bf KrishnaCam 50K} \\ 
& \textit{mini}-INet & INet & iNat & \textit{mini}-INet & INet & iNat & \textit{mini}-INet & INet & iNat & \textit{mini}-INet & INet & iNat \\ \midrule
SimCLR MemStoryboard & 35.02 & 20.72 & 5.65 & 39.58 & 24.78 & 7.77 & 33.72 & 20.13 & 5.64 & 36.36 & 22.75 & 6.10 \\
SimSiam MemStoryboard & 36.72 & 22.99 & 6.66 & 41.32 & 26.37 & 9.85 & 33.78 & 21.38 & 6.51 & 35.20 & 22.75 & 6.71 \\
Supcon Only & 34.62 & 21.04 & 6.62 & 39.08 & 24.92 & 8.19 & 31.68 & 21.29 & 5.86 & 34.92 & 23.07 & 6.56 \\
\bottomrule
\end{tabular}
}
\caption{Results on Memory Storyboard with only the temporal contrastive Loss.}
\label{tab:supcon_only}
\end{table}

\subsection{Memory Storyboard with Multiple Gradient Steps per Batch}
Using multiple gradient steps for each batch is a widely used technique in online continual learning~\cite{madaan2021representational}. We investigate the performance of Memory Storyboard when we take multiple gradient steps on each batch. Results are shown in Table~\ref{tab:multi_grad}. We observe that using multiple gradient steps (2 or 4) produces a sizable improvement on the ImageNet readout evaluation on KrishnaCam but not on the other benchmarks. We also observed that the improvement of multiple gradient steps is a lot smaller on SAYCam (sometimes even harming the performance), presumably due to the fact that SAYCam is a much larger training dataset than KrishnaCam and streaming learning without multiple gradient steps is sufficient for the model to capture a wide range of visual concepts.
\begin{table}[H]
    \vspace{-0.25in}
\centering
\resizebox{0.99\textwidth}{!}{
\begin{tabular}{l|c|ccc|ccc|ccc|ccc}
\toprule
\multirow{2}{*}{\bf Method} & \bf Grad & \multicolumn{3}{c|}{\bf SAYCam 10K} & \multicolumn{3}{c|}{\bf SAYCam 50K} & \multicolumn{3}{c|}{\bf KrishnaCam 10K} & \multicolumn{3}{c}{\bf KrishnaCam 50K} \\ 
& \bf Steps & \textit{mini}-INet & INet & iNat & \textit{mini}-INet & INet & iNat & \textit{mini}-INet & INet & iNat & \textit{mini}-INet & INet & iNat \\ \midrule
SimCLR MemStoryboard & 1 & 35.02 & 20.72 & 5.65 & 39.58 & 24.78 & 7.77 & 33.72 & 20.13 & 5.64 & 36.36 & 22.75 & 6.10 \\
SimCLR MemStoryboard & 2 & 34.56 & 23.47 & 5.66 & 38.44 & 26.49 & 7.67 & 33.00 & 23.86 & 5.45 & 35.30 & 26.36 & 6.38 \\
SimCLR MemStoryboard & 4 & 33.26 & 22.08 & 4.21 & 35.78 & 25.64 & 5.96 & 32.40 & 23.19 & 5.61 & 35.34 & 26.11 & 6.38 \\
SimSiam MemStoryboard & 1 & 36.72 & 22.99 & 6.66 & 41.32 & 26.37 & 9.85 & 33.78 & 21.38 & 6.51 & 35.20 & 22.75 & 6.71 \\
SimSiam MemStoryboard & 2 & 37.04 & 26.89 & 7.13 & 40.20 & 30.29 & 9.27 & 33.86 & 25.22 & 6.27 & 35.66 & 26.21 & 6.85 \\
SimSiam MemStoryboard & 4 & 35.02 & 22.88 & 6.73 & 36.30 & 25.21 & 7.36 & 33.44 & 24.34 & 6.44 & 34.98 & 25.73 & 6.62 \\
\bottomrule
\end{tabular}
}
\caption{Results on Memory Storyboard with different number of gradient update steps per batch.}
\label{tab:multi_grad}
\end{table}

\subsection{Memory Storyboard with Class-Balanced Buffer}
Inspired by other methods with use smart memory storage policies~\citep{yu2023scale, purushwalkam2022challenges}, we investigate the performance of Memory Storyboard with a class-balanced memory. When we attempt to add a new data point to the long-term memory that is already full, we randomly remove one of the data points from the class with the most samples in the memory. Results are shown in Table~\ref{tab:bal_buffer}. We observe that using the class-balanced memory produces mild improvements over the reservoir sampling baseline, though results on specific runs are mixed. We think that the memory storyboard method should work well with many different buffer sampling strategies, and advancements in buffer sampling strategies are orthogonal to the contribution of this work.
\begin{table}[H]
\centering
\resizebox{0.99\textwidth}{!}{
\begin{tabular}{l|c|ccc|ccc|ccc|ccc}
\toprule
\multirow{2}{*}{\bf Base SSL Method} & 
\multirow{2}{*}{\bf Bal. Buffer} & \multicolumn{3}{c|}{\bf SAYCam 10K} & \multicolumn{3}{c|}{\bf SAYCam 50K} & \multicolumn{3}{c|}{\bf KrishnaCam 10K} & \multicolumn{3}{c}{\bf KrishnaCam 50K} \\ 
& & \textit{mini}-INet & INet & iNat & \textit{mini}-INet & INet & iNat & \textit{mini}-INet & INet & iNat & \textit{mini}-INet & INet & iNat \\ \midrule
SimCLR & \xmark & 35.02 & 20.72 & 5.65 & 39.58 & 24.78 & 7.77 & 33.72 & 20.13 & 5.64 & 36.36 & 22.75 & 6.10 \\
SimCLR & \cmark & 34.04 & 22.21 & 6.41 & 38.58 & 26.99 & 7.58 & 31.92 & 21.28 & 5.70 & 34.66 & 23.64 & 5.78 \\ \midrule
SimSiam & \xmark & 36.72 & 22.99 & 6.66 & 41.32 & 26.37 & 9.85 & 33.78 & 21.38 & 6.51 & 35.20 & 22.75 & 6.71 \\
SimSiam & \cmark & 34.66 & 23.02 & 6.33 & 37.20 & 24.69 & 7.37 & 33.86 & 22.67 & 6.32 & 36.76 & 25.61 & 7.09 \\
\bottomrule
\end{tabular}
}
\caption{Results on Memory Storyboard with class-balanced buffer. 
}
\label{tab:bal_buffer}
\end{table}

\subsection{Additional Qualitative Results}
\begin{figure*}[t]
    \centering
    \includegraphics[width=0.95\linewidth,trim={0 0 0 0},clip]{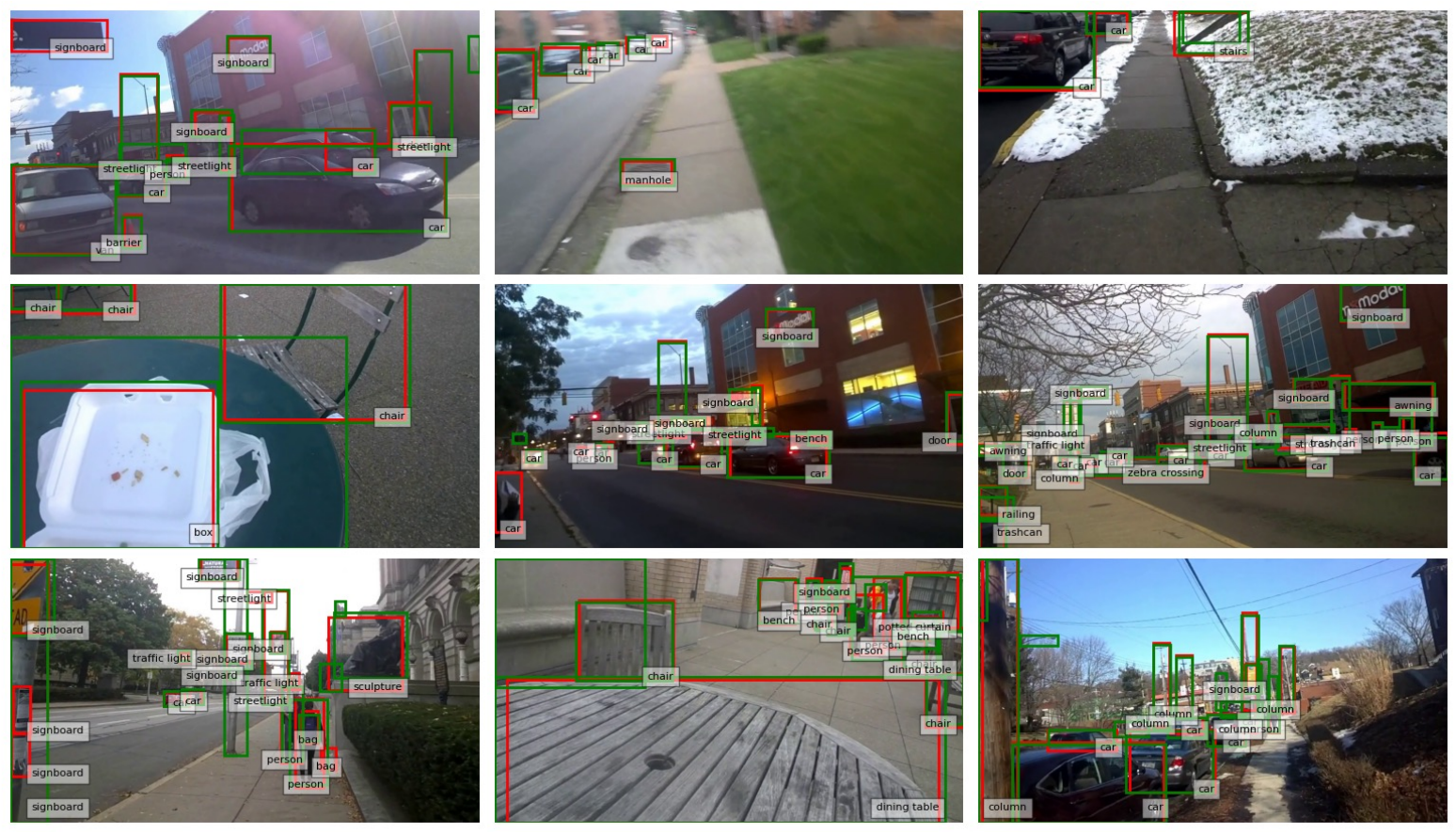}
    \caption{
    \textbf{Visualization of object detection results on the OAK validation set.} The Memory Storyboard model is trained on KrishnaCam and fine-tuned on the OAK training set. Red boxes show the predictions and the green boxes are ground truth bounding boxes.}
    \label{fig:oak_detection_vis}
\end{figure*}

\paragraph{OAK Object Detection Results.} We visualize the object detection results produced by Memory Storyboard when fine-tuned on the OAK dataset~\citep{wang2021wanderlust} in Figure~\ref{fig:oak_detection_vis}. We observe that the fine-tuned model can successfully detect objects in cluttered environments. The results show that the representations learned by Memory Storyboard can be effectively transferred to downstream tasks which requires more fine-grained features.

\begin{figure*}[t]
    \centering
    \includegraphics[width=0.99\linewidth,trim={0.5cm 0 0 0},clip]{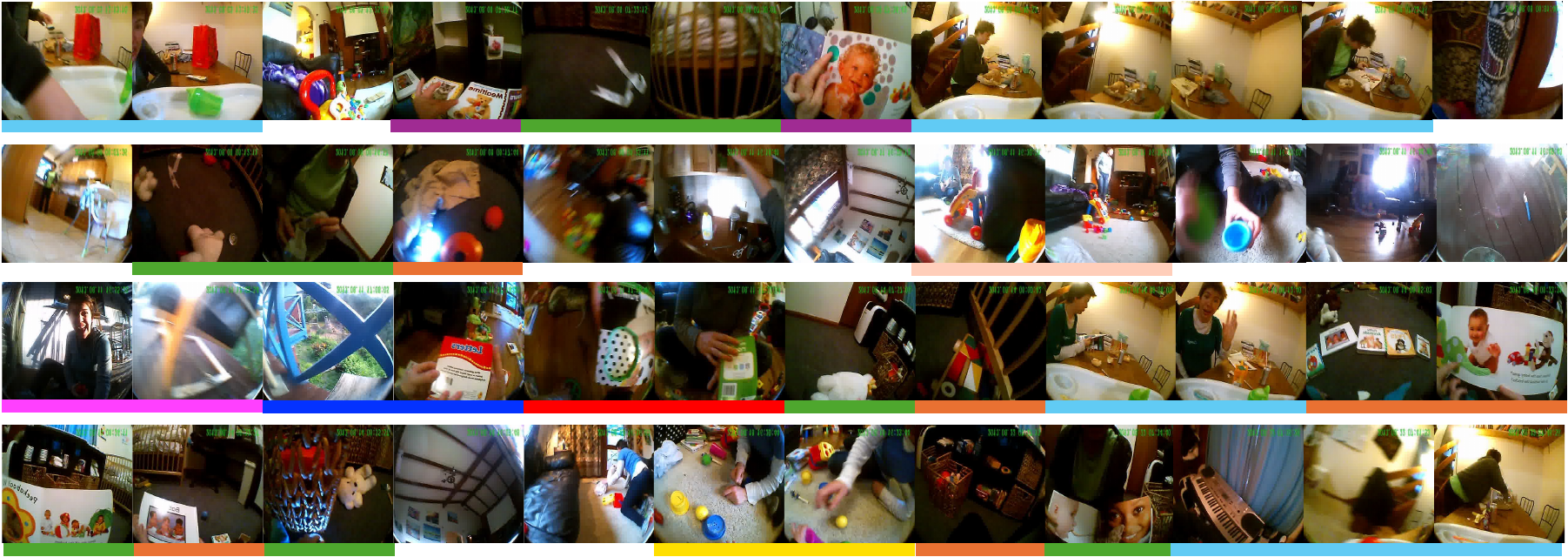}
    \caption{
    \textbf{Visualization of label merging by Memory Storyboard on SAYCam.} Each image represents a temporal segment; segments sharing the same color bar have been merged. Memory Storyboard successfully groups semantically similar scenes—e.g., segments marked with the light blue bar are all associated with the dining table.
    }
    \label{fig:saycam_merge_vis}
\end{figure*}

\paragraph{Label Merging Results.} We visualize the class labels produced by the label merging mechanism in Memory Storyboard in Figure \ref{fig:saycam_merge_vis}. We observe that Memory Storyboard can successfully group semantically similar scenes together, which helps improve the representation learning performance.

\begin{figure*}[t]
    \centering
    \includegraphics[width=0.99\linewidth,trim={0.5cm 0 0 0},clip]{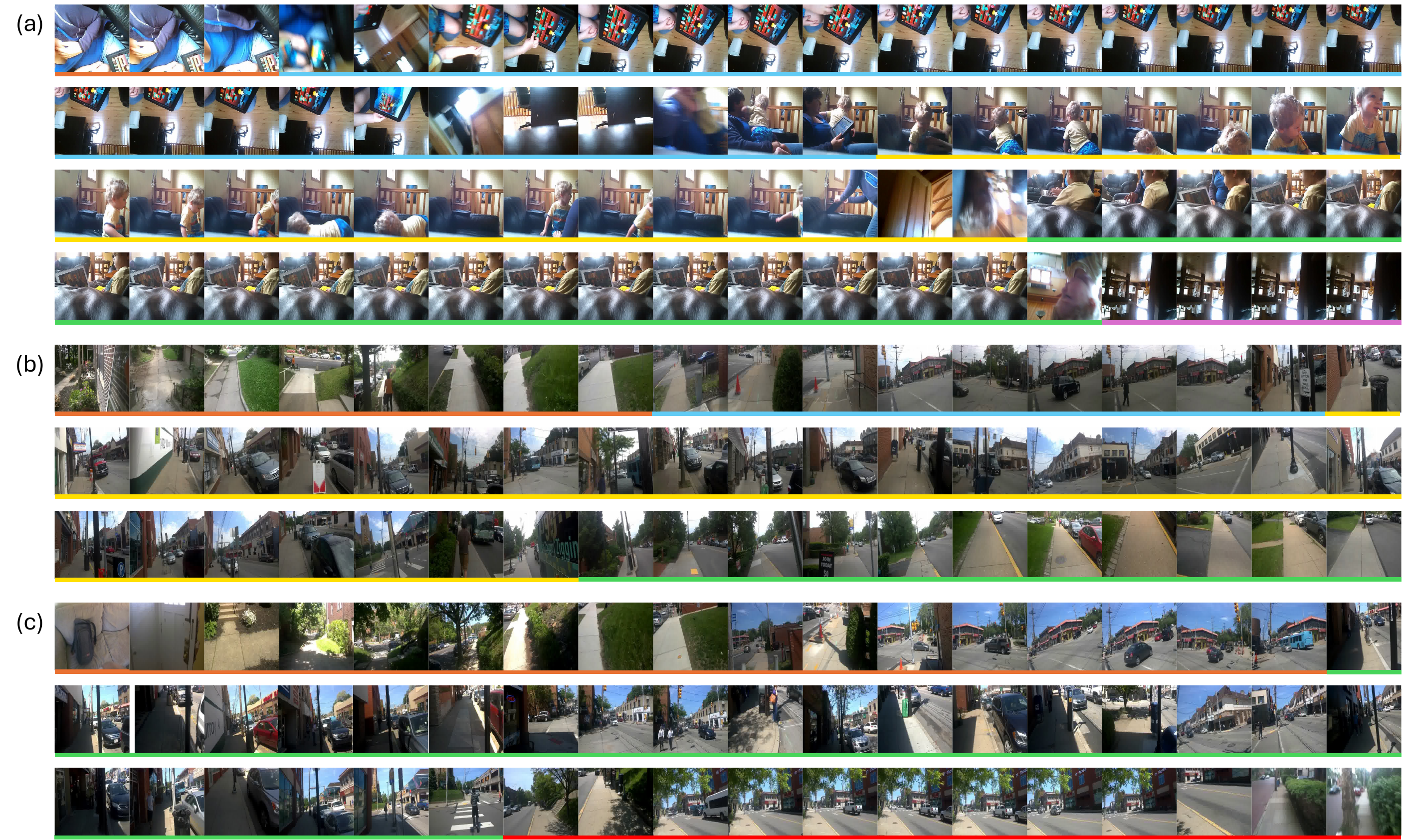}
    \caption{
    \textbf{Visualization of the temporal segments produced by randomly initialized models on (a) SAYCam (b)(c) KrishnaCam.} The images are the same as the ones in Figure~\ref{fig:saycam_seg_vis}. We observe that Memory Storyboard training enables to model to capture more intricate transitions between scenes.}
    \label{fig:saycam_seg_vis_scratch}
\end{figure*}

\paragraph{Temporal Segmentation by Randomly Initialized Models.} We visualize the temporal segments produced by randomly initialized models in Figure~\ref{fig:saycam_seg_vis_scratch}. By comparing to Figure~\ref{fig:saycam_seg_vis}, we observe that randomly initialized models fail to capture intricate transitions between scenes and cannot create accurate temporal segments, while Memory Storyboard training enables the model to learn better image representations to capture more intricate scene transitions.

\section{Optimal Batch Composition for SimCLR}

\begin{figure*}[t]
\centering\includegraphics[width=0.99\textwidth]{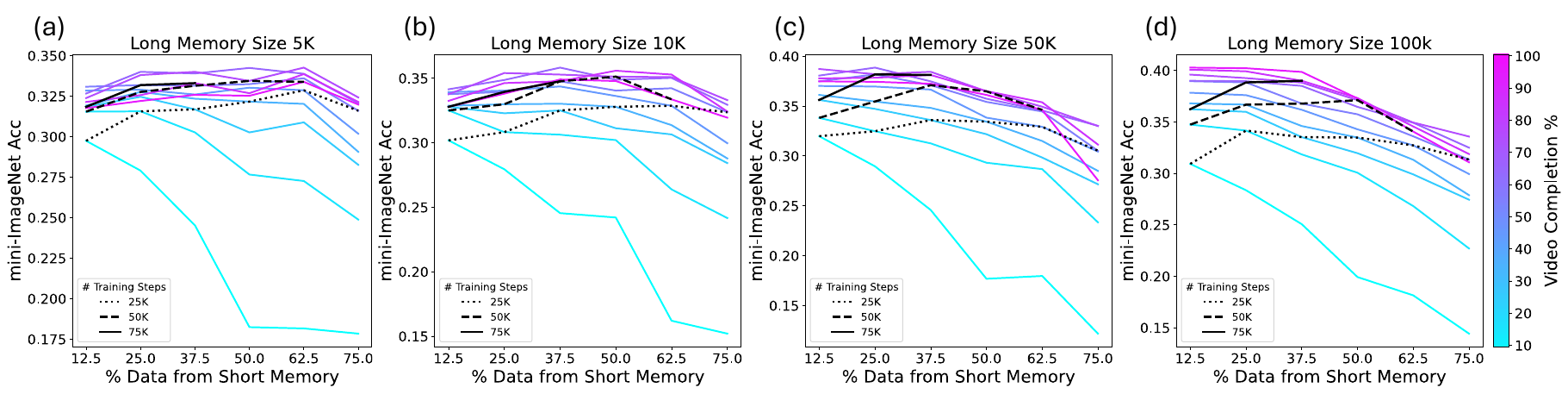}
    \caption{
    \textbf{SimCLR model performance on SAYCam with different long-term memory sizes} (5k, 10k, 50k, and 100k) \textbf{and varying training batch compositions} (12.5\% -- 75.0\% from $M_{short}$) using SVM readout. Each colored line represents the performance of different training batch compositions when \textbf{the model has seen the same amount of data} from the stream. Each black line represents the performance of different training batch compositions when the model has taken \textbf{the same number of gradient updates}.}
    \label{fig:batch_buffer_simclr}
\end{figure*}

\begin{figure*}[t]
\centering\includegraphics[width=0.99\textwidth]{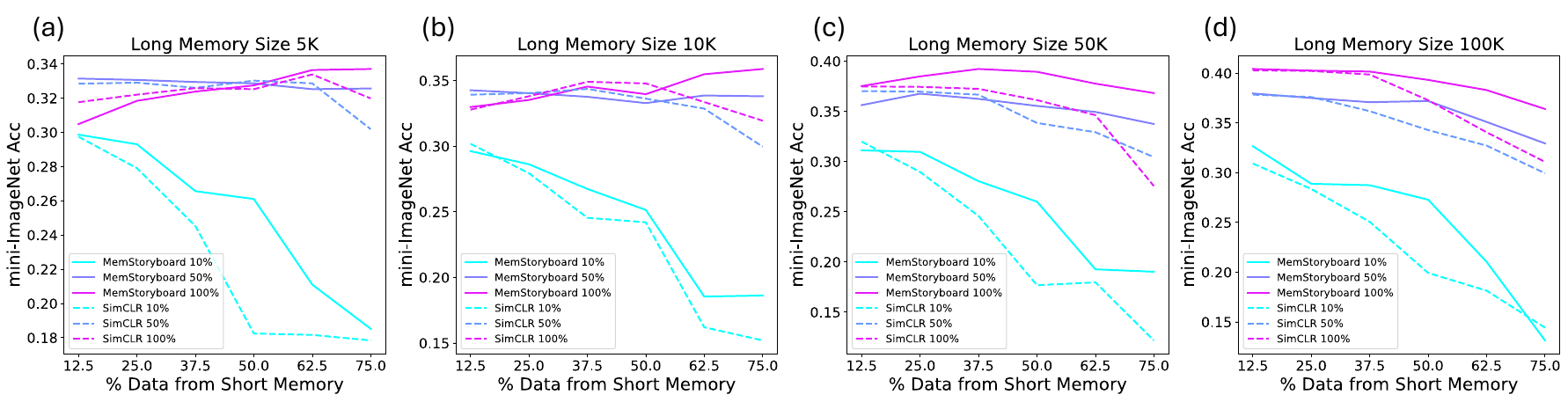}
    \caption{
    \textbf{Comparison of Memory Storyboard (solid lines) and SimCLR (dashed lines) model performance} on SAYCam using SVM readout, controlling the amount of data the model has seen from the stream.
    }\label{fig:video_completion_compare}
\end{figure*}

We replicate the experiments in Figure~\ref{fig:batch_buffer} on SimCLR models with two-tier memory, and plot the results in Figure~\ref{fig:batch_buffer_simclr}. We observe that the analysis and the conclusions of section~\ref{sec:batch_buffer} still hold: when we have a large memory, we either prefer balanced training batches (with a fixed amount of computation) or a bigger batch from long-term memory (with a fixed amount of data); when we can only afford a small memory, we prefer a smaller batch from long-term memory. We also want to note that the SVM readout results start to go down towards the end of the streaming training in SimCLR experiments more often than Memory Storyboard experiments, suggesting the better scalability of Memory Storyboard to larger-scale streaming training.

These results demonstrate that the analysis and observations in section~\ref{sec:batch_buffer} regarding the optimal batch composition for streaming SSL training under different memory and compute constraints are general, and apply to standard SSL methods in addition to Memory Storyboard.

\section{More Comprehensive Comparison between Memory Storyboard and SimCLR}
With the experiment results in Figure~\ref{fig:batch_buffer} and Figure~\ref{fig:batch_buffer_simclr}, we provide a more comprehensive comparison between Memory Storyboard and SimCLR performance under different memory constraints and batch compositions in Figure~\ref{fig:video_completion_compare}. We observe that Memory Storyboard outperforms SimCLR under the same amount of seen data, across a wide range of memory sizes and batch compositions. In particular, we note that Memory Storyboard significantly outperforms SimCLR when we sample more data from $M_{short}$ (towards the right side of the $x$-axis). This results in higher optimal performance when the memory size is small, where a larger batch from $M_{short}$ is needed to prevent overfitting on the long-term memory for better performance. We argue that, with temporal segmentation and the temporal contrastive loss, Memory Storyboard is able to provide better memory efficiency and also alleviate the temporal correlation issue suffered by SimCLR when we sample a large batch from the short-term memory.

\end{document}